\documentclass[runningheads]{llncs}

\usepackage{eccv}

\usepackage{eccvabbrv}

\usepackage{graphicx}
\usepackage{booktabs}
\usepackage{pifont}
\usepackage[table]{xcolor}
\usepackage{multirow}
\usepackage{wrapfig}
\usepackage{tabularx}
\usepackage{stackengine}
\usepackage{makecell}
\usepackage{enumitem}
\usepackage[accsupp]{axessibility}  

\usepackage{hyperref}

\usepackage{orcidlink}
\usepackage[table]{xcolor}

\newcommand{\modelname}{SafeGuard}
\newcommand{\datasetsname}{SafeVid}
 
\newcommand{\cmark}{\textcolor{green!70!black}{\ding{51}}}

\newcommand{\xmark}{\textcolor{red!90!black}{\ding{55}}}

\newcommand{\best}[1]{\textbf{#1}}
\newcommand{\second}[1]{\underline{#1}}
\definecolor{rose}{RGB}{235,80,160}

\begin{document}

\title{
\makebox[\textwidth][c]{
\parbox{1.0\textwidth}{
\centering
\raisebox{-0.55ex}{
\includegraphics[height=1.2em]{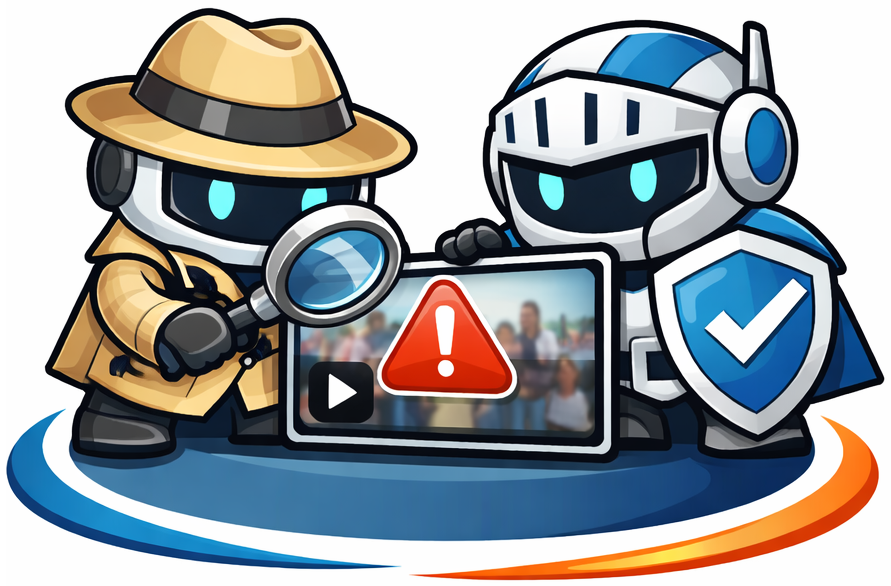}
}
\hspace{-0.9em}
SafeGuard: A Multi-Agent Perception-Reasoning Framework for Social-Risk AI-Generated Video Detection
}
}
}

\titlerunning{\modelname}

\newcommand{\equalcontrib}{\textsuperscript{*}} \newcommand{\corrauth}{\textsuperscript{\ensuremath{\dagger}}}
\renewcommand{\thefootnote}{}
\author{ Wenlin Wu\inst{1}\equalcontrib\orcidlink{0009-0007-1389-5877} \and 
Sheng Zhou\inst{2}\equalcontrib\orcidlink{0009-0007-4215-5464} \and 
Peipei Song\inst{1}\orcidlink{0000-0001-6764-3375} \and Wenhao Wang\inst{3}\orcidlink{0000-0001-8727-1572} \and Junbin Xiao\inst{1}\corrauth \and 
Xun Yang\inst{1}\corrauth\orcidlink{0000-0003-0201-1638} 
}

\footnotetext{${}^*$ Equal contribution; ${}^{\dagger}$ Corresponding authors: Xun Yang and Junbin Xiao.}

\authorrunning{W.~Wu et al.}

\institute{University of Science and Technology of China \and King Abdullah University of Science and Technology \and 
Vast Intelligence Lab\\ \email{wenlin097@mail.ustc.edu.cn, \{xyang21,junbinxiao\}@ustc.edu.cn}}

\maketitle

\begin{abstract}
As video generation paradigms evolve from localized manipulation to full-scene synthesis, AI-generated video detection becomes increasingly challenging, as forgeries exhibit coherent global structure and high perceptual realism. However, existing benchmarks are biased toward perceptual fidelity and primarily evaluate detectors based on perceptual artifacts, providing limited coverage of scenarios that require reasoning about violations of physical laws, structural coherence, or social logic. This dataset bias shapes current approaches and results in a \emph{Perception–Reasoning Gap}: artifact-centric models capture low-level statistical irregularities yet lack semantic inference, whereas vision-language models perform semantic reasoning but remain insensitive to fine-grained forensic cues. To bridge this gap, we propose \emph{\modelname}, a multi-agent framework that enables collaborative specialization between forensic perception and semantic reasoning. A hierarchical perceptual solver extracts fine-grained forensic evidence, while a self-reflective verifier enforces consistency between semantic inference and physical plausibility, forming an interpretable evidence chain. To support evaluation, we introduce \emph{\datasetsname}, a novel AI-generated video detection benchmark comprising 20K videos spanning 10 social risk categories, designed to evaluate physical plausibility, structural consistency, and the rationality of social behaviors. Extensive experiments demonstrate the generalization of \emph{\modelname}, improving accuracy on SafeVid by +18.7\% and consistently outperforming prior methods across four public benchmarks. The code and dataset are publicly available at {\color{rose}\url{https://github.com/williamw99/SafeGuard}}.

\end{abstract}

\begin{figure}[t!] 
    \centering 
    \includegraphics[width=1.0\textwidth]{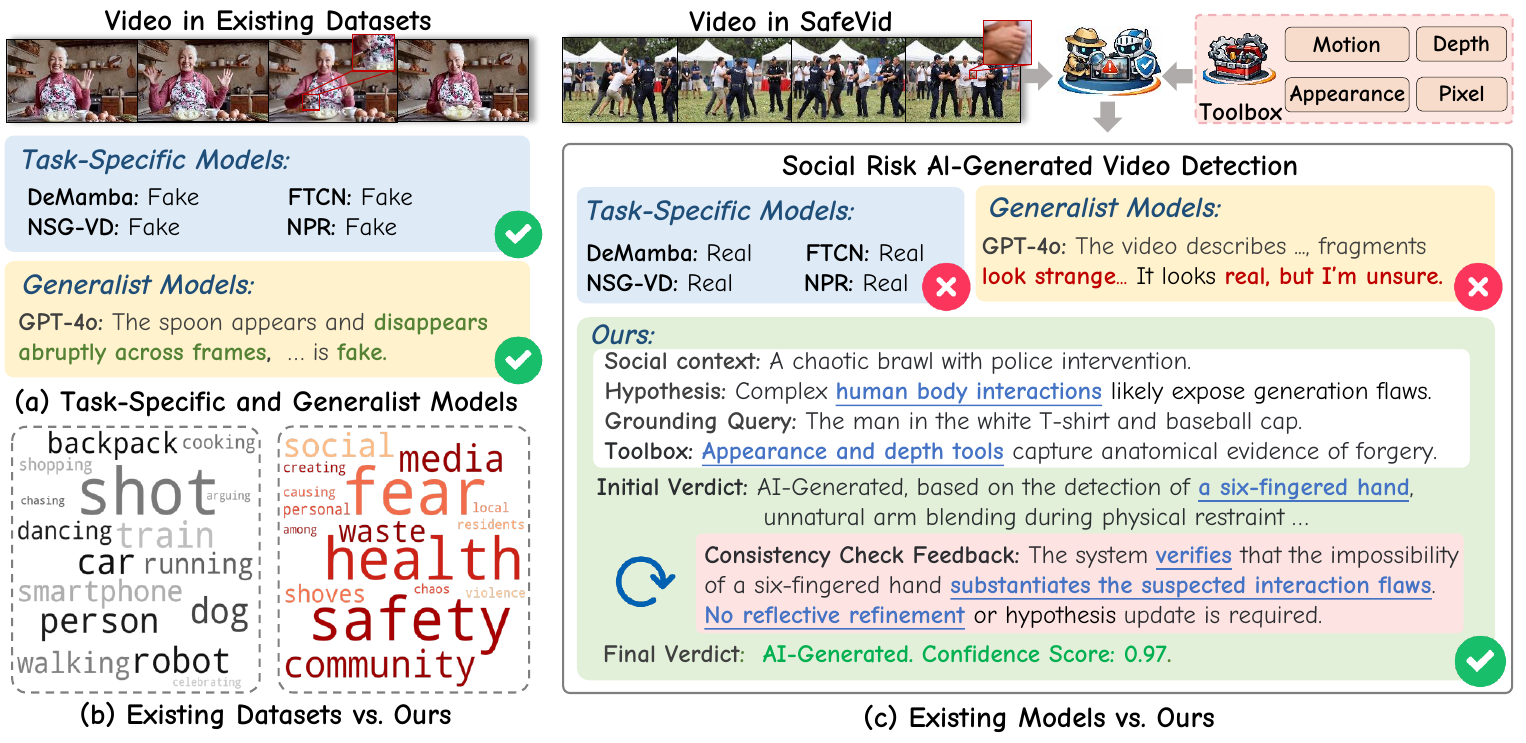} 
    \caption{\textbf{The illustration of social-risk AI-generated video detection.} Our framework integrates perceptual forensic evidence with semantic reasoning.
    }
    \label{fig:overall}
\end{figure}
\section{Introduction}
\label{sec:intro}
Recent advances in video generation have shifted synthesis from localized manipulation (\eg, facial editing~\cite{yao2021latent}) to full-scene generation. Modern systems, including Sora~\cite{openai_sora_system_card_2024}, Pika~\cite{pika}, Wan~\cite{wan2025wan},  Nano Banana Pro~\cite{comanici2025gemini}, and controllable generation~\cite{chengzhijing2026moca, ling2026everybodydance}, can produce globally coherent and semantically consistent videos that closely resemble authentic footage. Unlike early deepfake research~\cite{Identity_deepfake,lin2024detecting, wang2025scalinglawsdeepfakedetection}, which focused on identity manipulation, current generative models can synthesize complete events, including misinformation, violent incidents, and politically sensitive scenarios. As synthetic videos grow in semantic complexity and social impact, detection approaches extend beyond low-level artifact analysis to evaluate event-level authenticity and {contextual plausibility~\cite{wang2026affordbot}.} However, existing AI-generated video detection {benchmarks~\cite{wang2026videoufo,chen2024demambaaigeneratedvideodetection,mmdetnips2024,ni2025genvidbenchchallengingbenchmarkdetecting,wang2024vidprom}} (\eg, GenVideo~\cite{chen2024demambaaigeneratedvideodetection}, DVF~\cite{mmdetnips2024}, and GenVidBench~\cite{ni2025genvidbenchchallengingbenchmarkdetecting}) emphasize visual realism in benign scenarios, such as daily activities or natural scenes, as illustrated in Fig.~\ref{fig:overall} (a), while largely underrepresenting high-risk semantic contexts. This limited coverage creates a distribution gap between benchmark data and real-world social-risk environments, limiting the generalization of detection systems in safety-critical settings. These limitations motivate benchmarks and detection frameworks tailored to social-risk AI-generated videos, enabling semantically grounded and context-aware authenticity assessment.

However, existing AI-generated video detection methods face a fundamental \emph{Perception–Reasoning Gap}: task-specific detectors excel at extracting low-level forensic cues but struggle to assess physical feasibility or social-contextual consistency, resulting in limited robustness and interpretability. In contrast, large vision–language models, such as Qwen3-VL~\cite{Qwen3-VL}, InternVL3.5~\cite{wang2025internvl3}, and GPT-4o~\cite{gpt-4o}, provide strong semantic reasoning yet remain insensitive to subtle perceptual evidence. As illustrated in Fig.~\ref{fig:overall}~(c), this limitation leads to critical failures in social-risk scenarios.
This tension reflects an inherent trade-off: preserving fine-grained perceptual evidence for forensic analysis often conflicts with the abstraction required for semantic reasoning.

Motivated by this challenge, we propose \textbf{\modelname}, a multi-agent framework that combines the fine-grained evidence extraction capabilities of low-level visual forensic models with the high-level semantic reasoning abilities of large vision-language models. The framework formulates detection as a closed-loop, evidence-driven reasoning process. Specifically, it consists of two complementary modules: a \emph{Hierarchical Perceptual Solver} and a \emph{Self-Reflective Verifier}. The Solver adopts a coarse-to-fine strategy to localize critical regions and extract explicit forensic cues using lightweight specialized models. The Verifier then evaluates semantic–perceptual consistency by cross-validating the extracted evidence against contextual reasoning. As illustrated in Fig.~\ref{fig:overall}~(c), when analyzing a complex social event such as a chaotic brawl, the Solver first formulates a semantic hypothesis and deploys targeted tools to gather concrete forensic evidence (\eg, detecting a six-fingered hand). The verifier then leverages this physical anomaly as evidence to validate the hypothesis and reach the current verdict. Through this iterative mutual grounding process, \modelname\ reconciles low-level forensic sensitivity with high-level semantic reasoning, producing interpretable decisions supported by an explicit chain of evidence.

To validate our framework, we construct a social-risk AI-generated video detection benchmark, \textbf{\datasetsname}, consisting of 20K videos with 10K authentic and 10K AI-generated samples. The videos are produced by 21 state-of-the-art generative models and cover 10 social risk categories.
\datasetsname\ is designed to evaluate detection robustness under high-risk semantic conditions and to encourage generalization beyond visually benign scenarios, as illustrated in Fig.~\ref{fig:overall}.
Extensive experiments demonstrate that \modelname\ consistently outperforms state-of-the-art models on \datasetsname\ and four existing public benchmarks, including LOKI~\cite{ye2024loki}, GenVideo~\cite{chen2024demambaaigeneratedvideodetection}, GenVidBench~\cite{ni2025genvidbenchchallengingbenchmarkdetecting}, and DVF~\cite{mmdetnips2024}. Different from prior training-based detection methods that require task-specific fine-tuning on target datasets, our overall framework requires no end-to-end fine-tuning for the video detection task. Notably, our adopted lightweight forensic tools utilize publicly available GenVideo-finetuned weights, and our ablation studies confirm these optimized cues yield non-trivial performance gains in social-risk AI-generated video detection. Beyond superior detection accuracy, our framework provides a transparent reasoning process that links semantic suspicion, forensic cue extraction, and logical deduction, enabling interpretable assessment of socially sensitive risks.


In summary, the main contributions are as follows:

\begin{sloppypar}
\begin{itemize}[nosep,leftmargin=*]
\item We introduce the task of social-risk AI-generated video detection, requiring models to jointly assess physical plausibility, identity consistency, and social-logic coherence beyond traditional artifact-based inspection. 
\item We propose \modelname, a multi-agent framework that integrates fine-grained forensic cue extraction via a Hierarchical Perceptual Solver with semantic–physical consistency verification through a Self-Reflective Verifier, bridging the perception-reasoning divide to jointly analyze semantic falsehoods and visual forensic traces in AI-generated video detection. 
\item We construct \datasetsname, a benchmark of 20K videos generated by 21 state-of-the-art models across 10 socially critical risk categories, curated to evaluate detection robustness in semantically complex and socially sensitive scenarios.
\item Extensive experiments on \datasetsname\ and four public benchmarks demonstrate the effectiveness and generalization capability of \modelname, which achieves strong detection performance across diverse social-risk scenarios.
\end{itemize}
\end{sloppypar}

\section{Related Works}
\subsection{AI-Generated Video Detection Methods and Datasets} AI-generated video detection aims to distinguish synthetic videos from authentic content, and it inherently intersects with vision-language comprehension and reasoning paradigms~\cite{xiao2021next, chinchure2025spotlight, zhou2025scene, yang2026towards, li2025videochatr1,yangvmr, Deconfounded}. With the evolution of generative models, the task has progressed from low-level artifact inspection to jointly modeling perceptual cues and high-level semantic consistency. Existing approaches can be categorized as: \textbf{(1) Training-based methods.} These methods leverage supervised or instruction-tuned learning to capture discriminative forgery evidence. Early models focus on appearance, motion, or geometric inconsistencies, \eg, DeMamba~\cite{chen2024demambaaigeneratedvideodetection} and DeCoF~\cite{ma2024decofgeneratedvideodetection} model pixel- and flow-level artifacts, while WaveRep~\cite{corvi2025seeing}, ReStraV~\cite{interno2025aigeneratedvideodetectionperceptual}, and NSG-VD~\cite{zhang2025physics} incorporate physical or frequency-domain cues. Traditional AI-generated image detectors~\cite{ftcn,npr} and deepfake methods~\cite{coccomini2024mintime,xu2023tall,ni2022expanding} follow a similar paradigm, exploiting texture, blending, or warping artifacts. Recent training-based approaches~\cite{mmdetnips2024,wen2025busterxmllmpoweredaigeneratedvideo,li2025skyra} extend this with semantic modeling via large multimodal architectures, using multimodal alignment or instruction tuning to extract forgery evidence. However, they require substantial training and may underutilize fine-grained perceptual cues.   \textbf{(2) Training-free methods.} These approaches avoid task-specific optimization by leveraging external priors or foundation models. Physics-inspired methods like D3~\cite{zheng2025d3trainingfreeaigeneratedvideo} assess consistency with physical laws, whereas LLM-driven pipelines~\cite{gao2025david} such as LAVID~\cite{liu2025lavidagenticlvlmframework} frame detection as an agent-based reasoning process. Training-free methods reduce computational cost and improve interpretability but often sacrifice fine-grained perceptual grounding or semantic reasoning, limiting performance in complex scenarios.

While these methods advance detection capabilities, their evaluation is largely based on benchmarks covering generic or everyday scenarios. Existing datasets, such as GenVidBench~\cite{ni2025genvidbenchchallengingbenchmarkdetecting}, GenVideo~\cite{chen2024demambaaigeneratedvideodetection}, DVF~\cite{mmdetnips2024} and LOKI~\cite{ye2024loki}, primarily focus on general real-world content (\eg, landscapes and daily activities). Policy-oriented datasets like Video-SafetyBench~\cite{liu2025video} emphasize harmful content moderation but lack forensic-level annotations for real–synthetic discrimination. Consequently, current benchmarks provide limited coverage of socially sensitive and high-risk scenarios. To fill this gap, \datasetsname\ introduces an AI-generated video detection benchmark tailored to social safety, safety-critical events (\eg, protests and riots), enabling evaluation of detection frameworks that jointly model fine-grained perceptual cues and high-level semantic reasoning.

\subsection{Multi-Agent System for Visual Forensics} Multi-Agent Systems (MAS) address complex tasks by decomposing problem-solving into coordinated interactions among specialized agents, enabling scalable and modular {reasoning~\cite{hong2024metagpt,xu2026selfevolving}.} 
In visual forensics, MAS have been applied primarily to manipulation detection~\cite{lai2025agent4faceforgery, liang2025evidence, huang2025unishield}, such as face forgery and deepfake identification. 
Methods like UniShield~\cite{huang2025unishield} and AIFo~\cite{liang2025evidence} employ role-specialized agents for adversarial analysis, while Agent4FaceForgery~\cite{lai2025agent4faceforgery} uses agent societies to simulate forgery and evaluate detector robustness. 
These approaches excel at capturing fine-grained visual artifacts but typically lack semantic reasoning, limiting their interpretability in context-dependent manipulations.
MAS have also been explored in AI safety and multimodal risk assessment~\cite{asad2025reddebate}, using structured debate and adversarial argumentation to evaluate semantic risks and mitigate unsafe behaviors. 
While effective at high-level reasoning, they generally overlook subtle visual cues, reducing their applicability to perceptually grounded detection.
Overall, existing MAS approaches tend to focus on low-level visual forensics or high-level semantic reasoning, without a unified mechanism integrating both. 
Our framework demonstrates that orchestrating multi-agent collaboration with hierarchical perceptual and semantic reasoning can bridge this gap, enabling artifact detection and social event-level assessment.

\begin{figure*}[t]
    \centering
    \includegraphics[width=1.0\textwidth]{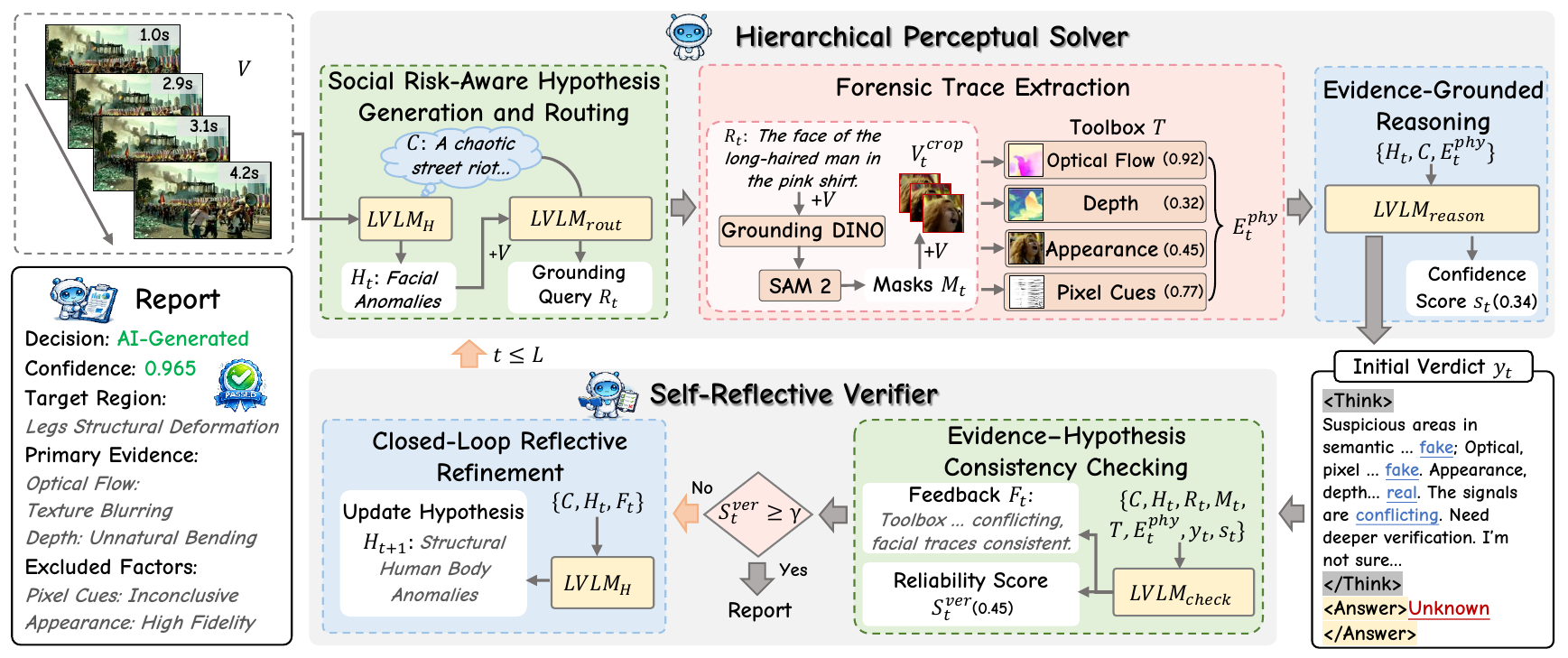}
    \caption{
    \textbf{The \modelname\ Framework.} \modelname\ combines a \textbf{Hierarchical Perceptual Solver} for coarse-to-fine trace extraction with a \textbf{Self-Reflective Verifier} that checks evidence–hypothesis consistency and performs closed-loop refinement.
    }
    \label{fig:scope_framework}
\end{figure*}

\section{The \modelname\ Framework}
\label{sec:method}
In this section, we propose an agent-centric framework, {\modelname}, which separates low-level perception and high-level reasoning into a \emph{Hierarchical Perceptual Solver} and a \emph{Self-Reflective Verifier}, respectively. The solver generates social risk-aware forgery hypotheses, extracts structured forensic evidence, and produces an initial verdict. The verifier audits this output for evidence–hypothesis consistency and performs reflective refinement when grounding is insufficient. This two-level design structures AI-generated video detection into semantic suspicion, evidence grounding, and logically validated decision-making.

\subsection{Hierarchical Perceptual Solver}

\subsubsection{Social Risk-Aware Hypothesis Generation and Routing} AI-generated video forgeries often concentrate in semantically and socially salient regions (\eg, human bodies or interactions), while static content like buildings is usually well-generated, as shown in Fig.~\ref{fig:scope_framework}. To focus analysis on high-risk areas, we introduce a social risk-aware module for hypothesis generation and routing. A large vision-language model (LVLM) summarizes scene dynamics into a global description $C$ (\eg, a chaotic street riot with intense interactions) and generates a forgery hypothesis $H_t$ specifying plausible artifact types (\eg, identity swapping, limb distortion, temporal tearing):
\begin{equation}
H_t = \text{LVLM}_{\text{H}}(C).
\end{equation}

This hypothesis then guides coarse spatial–temporal routing, identifying candidate regions $R_t$ for analysis while filtering out irrelevant background:
\begin{equation}
R_t = \text{LVLM}_{\text{rout}}(V, C, H_t).
\end{equation}

By coupling contextual abstraction with hypothesis-driven region selection, this module concentrates forensic scrutiny on semantically high-risk areas, improving detection accuracy and efficiency without sacrificing reliability.

\subsubsection{Forensic Trace Extraction}
\label{sec:forensic_trace}
High-level semantic analysis alone is insufficient for video authentication, as generation artifacts are often subtle, localized, and physically grounded. LVLMs may overlook fine-grained inconsistencies such as motion jitter, geometric distortions, or boundary noise. Hence, the solver incorporates a localized and physically sensitive mechanism to verify suspected manipulations.
Specifically, this module performs fine-grained forensic analysis on candidate regions identified by the preceding module.
We adopt a language-guided segmentation pipeline for precise region localization.
Given the grounding query $R_t$, frame-wise spatial grounding is conducted over the $N$ input frames of $V$.
Grounding DINO~\cite{liu2024grounding} is employed to predict a set of localized bounding boxes $\{B_t^{i}\}_{i=1}^{N}$.
Conditioned on these bounding boxes, SAM~2~\cite{ravi2024sam} generates the corresponding pixel-level masks
$M_t = \{M_t^{i}\}_{i=1}^{N}$. The masks $M_t$ are subsequently applied to the original video $V$ to isolate the target content, yielding a set of focused regions $V_t^{\text{crop}} = V \odot M_t$.
Structured forensic cues $E_t^{\text{phy}}$ are then derived via a task-specific toolbox $T$, where each tool assesses the authenticity of the cropped region from complementary physical perspectives.
\begin{equation}
\label{eq:ephy}
E_t^{\text{phy}}\!=\!
\Big\{
f_{\text{o}}(V_t^{\text{crop}}),
f_{\text{d}}(V_t^{\text{crop}}),
f_{\text{a}}(V_t^{\text{crop}}),
f_{\text{p}}(V_t^{\text{crop}})
\Big\},
\end{equation}
where
$f_{(\cdot)}(V_t^{\text{crop}})\in(0,1)$ represents the estimated probability that the region is real. Specifically, $f_{\text{o}}(\cdot)$ captures low-level motion information through optical flow to detect temporal inconsistencies, $f_{\text{d}}(\cdot)$ focuses on geometric cues derived from depth to identify physically implausible structures, $f_{\text{a}}(\cdot)$ encodes discriminative appearance features to ensure object-level identity consistency, and $f_{\text{p}}(\cdot)$ examines high-frequency pixel-level signals, such as boundary noise and subtle manipulation traces. By integrating these confidence scores, the system establishes an evidential representation that captures physical inconsistencies across motion, depth, appearance, and pixel-level cues.

\subsubsection{Evidence-Grounded Reasoning}
The presence of localized forensic cues $E_t^{\text{phy}}$ alone is insufficient for reliable video integrity assessment, as low-level irregularities may admit multiple semantic interpretations. In complex social scenarios, identical physical patterns could arise from either natural dynamics (\eg, rapid motion or occlusion) or generative manipulation. Consequently, decision-making requires reasoning that evaluates evidence under semantic and hypothesis constraints.
To this end, this stage performs hypothesis-conditioned reasoning by jointly modeling the extracted evidence $E_t^{\text{phy}}$, the global semantic context $C$, and the forgery hypothesis $H_t$. Rather than directly mapping cues to a categorical outcome, the model constructs a structured inference process that examines whether the observed physical signals are causally and logically consistent with the anomaly mechanism postulated in $H_t$. Formally, the reasoning module produces an initial verdict $y_t$ and a confidence score $s_t$:
\begin{equation}
(y_t, s_t) = \text{LVLM}_{\text{reason}}\!\left(H_t, C, E_t^{\text{phy}} \right),
\end{equation}
By integrating context, hypothesis, and localized forensic cues, this reasoning converts isolated low-level observations into a coherent forensic narrative (\eg, detecting temporal tearing inconsistent with motion dynamics). This ensures that conclusions are both logically grounded and physically traceable, providing the transparency and reliability required for high-stakes video forensics.

\subsection{Self-Reflective Verifier}
\subsubsection{Evidence–Hypothesis Consistency Checking}
Although the Hierarchical Perceptual Solver generates an initial verdict $y_t$ with a confidence score $s_t$, this preliminary result may still contain conflicts or {hallucinations~\cite{bai2024hallucination,fine2026pami}.} 
To ensure the reliability of the reasoning process, this stage functions as a rigorous gatekeeper. Its primary objective is to validate the coherence of the solver's entire execution trace before accepting the final verdict. Specifically, the verifier performs a multi-criteria audit that scrutinizes the plausibility of the hypothesis $H_t$ within the social risk context $C$, the precision of the segmentation masks $M_t$ in isolating target regions $R_{t}$, the appropriateness of the four task-specific tools $T = \{f_{\text{o}}, f_{\text{d}}, f_{\text{a}}, f_{\text{p}}\}$ generating $E_t^{\text{phy}}$ regarding the scene dynamics,
and the logical consistency between the extracted cues $E_t^{\text{phy}}$ and the initial verdict $y_t$. This verification yields a reliability score $S_t^{\text{ver}}$ together with feedback $F_t$:
\begin{equation}
(S_t^{\text{ver}},F_t) = \text{LVLM}_{\text{check}}(C, H_t, R_t, M_t, T, E_t^{\text{phy}}, y_t, s_t),
\end{equation}
where specific failure modes can be explicitly identified. When low reliability is detected, the verifier not only flags errors but also corrects reasoning results: unlike passive auditing, it acts as an active corrector by translating identified failure modes into natural-language feedback $F_t$ (\eg, ``the motion evidence contradicts the static background assumption''), which provides actionable guidance for correction. If the score falls below a critical reliability threshold (\ie, $S_t^{\text{ver}} < \gamma$), the system rejects the initial verdict and triggers the subsequent refinement phase.

\subsubsection{Closed-Loop Reflective Refinement}
When $S_t^{\text{ver}} < \gamma$, the system leverages the feedback $F_t$ to update the hypothesis from $H_t$ to $H_{t+1}$ and re-initiates the perception-reasoning cycle. This iteration continues until the verdict passes the consistency check or reaches the predefined cycle bound $L$.
\begin{equation}
H_{t+1} =
\begin{cases}
H_t, & \text{if } S_t^\text{ver} \ge \gamma \text{ or } t > L, \\
\text{LVLM}_{\text{H}}(C, H_t, F_t), & \text{otherwise}.
\end{cases}
\label{eq6}
\end{equation}
By integrating rigorous consistency verification with adaptive refinement, the proposed verifier ensures that the final reports are not only accurate but also explanatorily grounded and explicitly traceable to concrete forensic evidence, while preserving logical coherence. This reliability-aware design is especially crucial in high-stakes forensic contexts, where minimizing artifact-induced false positives is as important as ensuring precise detection.
\section{The \datasetsname\ Dataset}
\label{sec:datasets}

To facilitate research on social-safety-oriented video forensics, we construct a new benchmark named \datasetsname. In this section, we first introduce the safety-driven taxonomy. We then describe the procedures for real-world video collection and diversified synthetic video generation under multiple paradigms. Finally, we present statistical analysis and comparisons with existing related benchmarks.

\subsection{Datasets Construction}

\subsubsection{\datasetsname\ Taxonomy}
To cover representative social safety risks, we define a taxonomy of 10 fine-grained categories aligned with safety guidelines~\cite{zhao2025qwen3guard} and existing {benchmarks~\cite{wang2025tip,liu2025video,li2026unim, Zhou_2025_CVPR}.} These categories are \textit{Violence and Aggression}, \textit{Public Safety Incidents}, \textit{Corruption and Ethical Violations}, \textit{Radical Behavior and Terrorism}, \textit{Criminal Activities}, \textit{Misinformation and Propaganda}, \textit{Discrimination and Exclusion}, \textit{Public Health Risks}, \textit{Invasion of Privacy}, and \textit{Industrial and Environmental Hazards}. The taxonomy forms the semantic backbone of \datasetsname, guiding both real-world video collection and synthesized video generation.

\subsubsection{Real-World Video Collection}
To ensure diversity and realism, we collect videos from three public sources: YouTube~\cite{youtube_homepage_2025}, OpenVid-1M~\cite{nan2024openvid}, and UltraVideo~\cite{xue2025ultravideo}.
To align with the predefined taxonomy, we design a filtering strategy to filter out irrelevant videos. CLIP~\cite{radford2021learning} computes semantic similarity between candidate videos and taxonomy keywords, and videos are retained if the similarity exceeds a predefined threshold of {0.75}. This filtering approach mitigates noise introduced by keyword-based retrieval and ensures semantic consistency with the predefined public safety categories. Through this process, we obtain 10,178 authentic videos.

\begin{figure}[t!]
    \centering
        \centering
        \includegraphics[width=1.0\linewidth]{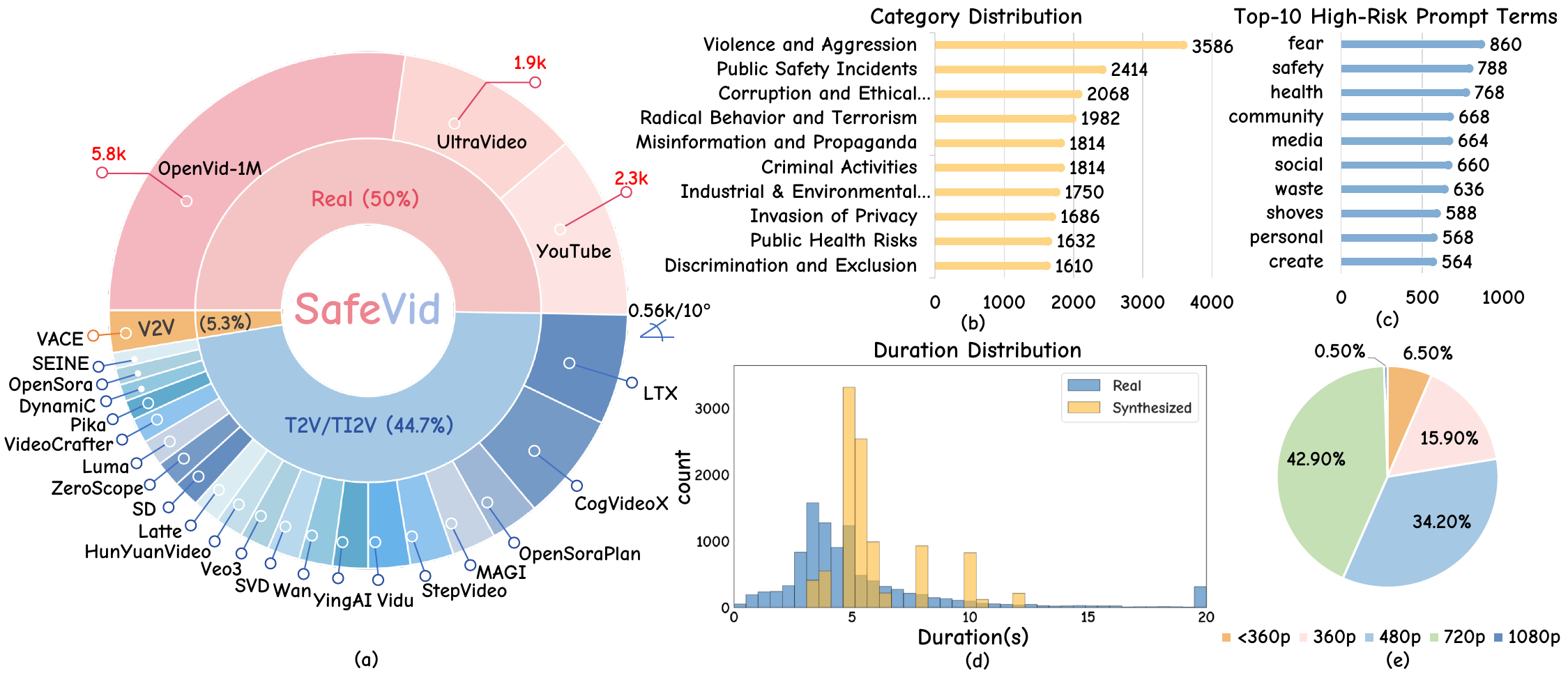}
        \caption{\textbf{\datasetsname\ statistics.} (a) Video count distribution. (b) Social risk category distribution. (c)  Top-10 high-risk terms in video generation prompts (including T2V and TI2V). (d) Video duration distribution. (e) Video resolution distribution.}
        \label{fig:dataset_stat}
    \hfill
\end{figure}

\subsubsection{Synthesized Video Generation}
To simulate an evolving social risk landscape, we construct synthetic data under two settings: 
\textbf{(1) Text-Driven Video Synthesis.} Given a safety keyword (\eg, traffic accident) and category (\eg, public safety incidents), we build scenario prompts and enrich them with scene attributes (\eg, lighting, motion, environment) using GPT-4o~\cite{gpt-4o}, forming a diverse prompt pool. Videos are then generated via two routes: \textcircled{1} \emph{Text-to-Video (T2V)}, rendered by multiple state-of-the-art models (\eg, Veo3~\cite{deepmind2025veo3modelcard}, Wan~\cite{wan2025wan}, CogVideoX~\cite{yang2025cogvideox}, HunYuanVideo~\cite{kong2024hunyuanvideo}); 
\textcircled{2} \emph{Text+Image-to-Video (TI2V)}, where anchor images produced by a text-to-image model (\eg, Qwen-Image~\cite{wu2025qwenimagetechnicalreport}) are animated by an image-to-video model (\eg, SVD~\cite{blattmann2023stablevideodiffusionscaling}). This process yields 6,473 T2V and 3,166 TI2V videos. \textbf{(2) Real-Video Editing.} To diversify manipulation patterns, we perform video-to-video editing on authentic videos. Semantic targets (\eg, identity or object replacement) are segmented using SAM 2~\cite{ravi2024sam}, and edited via a V2V model (\eg, VACE~\cite{jiang2025vace}) conditioned on the source video, mask, and instruction. This results in 539 edited videos with varied artifacts.

\begin{table*}[t]
\centering
\normalsize
\renewcommand\arraystretch{1.05}
\caption{\textbf{Comparison of AI-generated video detection datasets}. \textbf{Scenario}: \textit{General} denotes all open-domain scenes, such as cartoons and animations; \textit{Real-World} consists of videos resembling realistic, physically plausible environments; 
and \textit{Social Safety} focuses on social safety-oriented scenarios curated for high-risk or sensitive content.
\textbf{Tax.} indicates the fine-grained social safety taxonomy. 
}
\vspace{0.5mm}
\label{tab:video_dataset_comparison_abbr}
\resizebox{\linewidth}{!}{

\begin{tabular}{l c c c c c l c c}
\hline
\multirow{2}{*}{\textbf{Dataset}} &
\multirow{2}{*}{\textbf{Scenario}} &
\multicolumn{3}{c}{\textbf{Gen. Paradigm}} &
\multirow{2}{*}{\textbf{Tax.}} &
\multirow{2}{*}{\textbf{Video Source}} &
\multirow{2}{*}{\textbf{\#Mod.}} &
\multirow{2}{*}{\textbf{\#Vid.}} \\
\cline{3-5}
 & & \textbf{T2V} & \textbf{TI2V} & \textbf{V2V} & & & & \\
\hline

GVD~\cite{aigvdet} 
& General & \cmark & \cmark & \xmark & \xmark 
& VOS2, Got-10k & 8 & 22k \\

GVF~\cite{ma2024decofgeneratedvideodetection} 
& General & \cmark & \xmark & \xmark & \xmark 
& MSCD, MSR-VTT & 8 & 5k \\

GenVidDet~\cite{ji2024distinguishfakevideosunleashing} 
& General & \cmark & \cmark & \xmark & \xmark 
& InternVid, HD-VG-130M & 8 & \textbf{73k} \\

DVF~\cite{mmdetnips2024} 
& General & \cmark & \cmark & \xmark & \xmark 
& Youtube-8M, InternVid-10M & 8 & 4k \\

GenVideo~\cite{chen2024demambaaigeneratedvideodetection} 
& General & \cmark & \cmark & \xmark & \xmark 
& Kinetics-400, Youku-mPLUG, MSR-VTT & 20 & 19k \\

GenVidBench~\cite{ni2025genvidbenchchallengingbenchmarkdetecting} 
& General & \cmark & \cmark & \xmark & \xmark 
& Vript, HD-VG-130M & 8 & 69k \\

LOKI~\cite{ye2024loki} 
& General & \cmark & \xmark & \xmark & \xmark 
& - & 7 & 1k \\

VidForensic~\cite{liu2025lavidagenticlvlmframework} 
& General & \cmark & \xmark & \xmark & \xmark 
& PANDA-70M, Youtube & 8 & 2k \\

GenWorld~\cite{chen2025genworlddetectingaigeneratedrealworld} 
& Real-World & \cmark & \cmark & \xmark & \xmark 
& Kinetics-400, Nuscenes, RT-1, DL3DV-10K & 10 & 20k \\

GenBuster~\cite{wen2025busterxpp} 
& Real-World & \cmark & \xmark & \xmark & \xmark 
& OpenVid-1M & 12 & 4k \\

\hline
\rowcolor{gray!20}
\textbf{\datasetsname~(Ours)} 
& \textbf{Social Safety} 
& \textbf{\cmark} 
& \textbf{\cmark} 
& \textbf{\cmark} 
& \textbf{\cmark} 
& \textbf{YouTube, OpenVid-1M, UltraVideo} 
& \textbf{21} 
& {20k} \\

\hline
\end{tabular}
}
\end{table*}

\subsection{Dataset Analysis and Comparison}
\subsubsection{Statistics and Analysis} \datasetsname\ contains 20,356 videos, evenly divided into 10,178 real and 10,178 synthetic videos. Specifically, T2V/TI2V videos constitute 44.7\% and V2V samples 5.3\%, complementing the 50\% real portion. This balanced composition enables evaluation across both fully generative and real-video editing scenarios, enhancing artifact diversity and forensic difficulty.
The dataset is explicitly designed for social safety assessment. As shown in Fig.~\ref{fig:dataset_stat}~(b), \textit{Violence and Aggression} accounts for the largest share ($\sim$18\%), with the remaining categories evenly distributed, ensuring broad coverage of socially sensitive contexts.
The prompt statistics for video generation (including T2V and TI2V) in Fig.~\ref{fig:dataset_stat}~(c) further demonstrate the dataset’s safety-oriented design. High-frequency terms such as \textit{fear}, \textit{safety}, and \textit{health} indicate an alignment between the semantic distribution of generated content and real-world risk scenarios.

\subsubsection{Dataset Comparison}
As shown in Table~\ref{tab:video_dataset_comparison_abbr}, \datasetsname\ is the first AI-generated dataset dedicated to the {social safety} domain, focusing on safety-critical events rather than generic content.
Unlike prior datasets that predominantly rely on T2V synthesis (with limited TI2V synthesis and no V2V editing), \datasetsname\ encompasses T2V, TI2V, and V2V, enabling evaluation across both fully generative and real-video editing scenarios and increasing artifact diversity.
In addition, \datasetsname\ introduces a fine-grained safety taxonomy and incorporates 21 generative models to enhance source heterogeneity. The combination of multi-paradigm synthesis, safety-aware categorization, and diverse generation models establishes a challenging benchmark for social safety assessment.

\section{Experiments}

\subsection{Experimental Setup}

\subsubsection{Model Evaluation}
To assess the effectiveness and generalization of \modelname, we benchmark it against state-of-the-art approaches that reflect two paradigms in AI-generated video detection: \textbf{(1) Task-specific models.} These approaches are explicitly developed for AI-generated video detection and rely on supervision tailored to forensic discrimination. We include \textit{trainable} approaches spanning both image-level (NPR~\cite{npr}) and video-level (FTCN~\cite{ftcn}, TALL~\cite{xu2023tall}, XCLIP~\cite{ni2022expanding}, MINTIME~\cite{coccomini2024mintime}, DeMamba~\cite{chen2024demambaaigeneratedvideodetection}, ReStraV~\cite{interno2025aigeneratedvideodetectionperceptual}, NSG-VD~\cite{zhang2025physics}, BusterX++~\cite{wen2025busterxpp}, 
and Skyra-RL~\cite{li2025skyra}) modeling. We further evaluate the \textit{training-free} approach D3~\cite{zheng2025d3trainingfreeaigeneratedvideo} to assess artifact-level generalization without task-specific retraining. \textbf{(2) Generalist models.} To evaluate the performance of general-purpose large vision-language foundation models on AI-generated video detection, we benchmark these models under a zero-shot prompting setting. We consider advanced open-source models, including LLaVA-Video~\cite{zhang2024llavanextvideo},VideoLLaMA3~\cite{damonlpsg2025videollama3}, InternVL3.5~\cite{wang2025internvl3}, Qwen3-VL~\cite{Qwen3-VL}, and reasoning-oriented Video-R1~\cite{feng2025videor}, Video-RFT~\cite{wang2025videorft}, VideoChat-R1~\cite{li2025videochatr1}, as well as the proprietary model GPT-4o~\cite{gpt-4o}, and Gemini-2.5-Pro~\cite{comanici2025gemini}. All generalist models are evaluated using a unified prompt template to ensure a fair comparison. Detailed implementation settings, including hyperparameters and prompt templates, are provided in the Appendix.

\subsubsection{Datasets}
We evaluate all methods on our SafeVid dataset and four widely used public benchmarks: 
\texttt{GenVideo}~\cite{chen2024demambaaigeneratedvideodetection}, a large-scale benchmark for AI-generated video detection comprising 2,262k training videos and 18.5k testing videos; \texttt{DVF}~\cite{mmdetnips2024}, which includes 6.7k diffusion-generated videos produced by eight distinct diffusion-based generators; \texttt{LOKI}~\cite{ye2024loki}, a benchmark featuring 1.3k synthesized videos in diverse real-world contexts; and \texttt{GenVidBench}~\cite{ni2025genvidbenchchallengingbenchmarkdetecting}, containing 143k videos and designed to assess cross-model and cross-scenario generalization. For experiments, the training set is fixed and drawn from the \texttt{GenVideo}~\cite{chen2024demambaaigeneratedvideodetection} training set, \ie, $D_{\text{train}} = \{\texttt{GenVideo}_{\text{train}}\}$. The test set is partitioned into two complementary scenarios: \textbf{(1) In-Domain (ID)}, containing 4.6k video samples generated using the same sources and methods as in the training set; \textbf{(2) Out-of-Domain (OOD)}, containing 15.4k videos from completely unseen sources and generative models. Formally, the overall evaluation is defined as $D_{\text{test}} = D_{\text{\datasetsname}} \cup D_{\text{public}}$, where $D_{\text{\datasetsname}}=\{R_{\text{ID}}, S_{\text{ID}}, R_{\text{OOD}}, S_{\text{OOD}}\}$ ($R$ denotes real videos, $S$ denotes synthesized videos) and $D_{\text{public}}=\{\texttt{GenVideo}_{\text{test}}, \texttt{DVF}, \texttt{LOKI}, \allowbreak \texttt{GenVidBench}\}$. The data sources for \texttt{\datasetsname} splits are summarized in Table~\ref{tab:dataset_split}.

\begin{table}[t!]
    \centering
    \caption{
        \textbf{Dataset splits of \datasetsname.} 
        The dataset comprises \textbf{Real} and \textbf{Synthesized} videos 
        (\ie, videos generated or edited by T2V/TI2V/V2V models). 
    }
    \label{tab:dataset_split}
    \scriptsize
    \renewcommand{\arraystretch}{1.0}

    \newcolumntype{Y}{>{\centering\arraybackslash}X}

    \begin{tabularx}{\linewidth}{l| >{\hsize=0.85\hsize}Y | >{\hsize=1.15\hsize}Y }
    \toprule
    \textbf{Type} & \textbf{In-Domain (ID)} & \textbf{Out-of-Domain (OOD)} \\
    \midrule
    
    \textbf{Real} 
    & \texttt{YouTube} (2.3k)
    & \texttt{OpenVid-1M}, \texttt{UltraVideo} (7.7k) \\
    
    \midrule
    
    \textbf{Synthesized} 
    & \makecell{\texttt{SVD}, \texttt{OpenSora}, \texttt{DynamiC}, \\ 
                \texttt{Latte}, \texttt{ZeroScope}, \texttt{Pika}, \texttt{SD}, \\
                \texttt{SEINE}, \texttt{VideoCrafter}} (2.3k)
    & \makecell{\texttt{LTX}, \texttt{YingAI}, \texttt{StepVideo}, \texttt{Vidu}, \texttt{CogVideoX}, \\ 
                \texttt{OpenSoraPlan}, \texttt{HunyuanVideo}, \texttt{MAGI}, \texttt{VACE}, \\ 
                \texttt{Veo3}, \texttt{Wan}, \texttt{Luma}} (7.7k)\\
    
    \bottomrule
    \end{tabularx}
\end{table}

\subsubsection{Evaluation Metric}
Following prior work~\cite{li2025skyra}, we adopt Accuracy (\textbf{Acc}) and F1-score (\textbf{F1}) as our primary evaluation metrics, with F1 computed using \textit{Synthesized} videos as the positive class based on hard predictions. For task-specific trainable models, we additionally report Area Under the Curve (\textbf{AUC}), Average Precision (\textbf{AP}), and Equal Error Rate (\textbf{EER}) in the Appendix following~\cite{chen2024demambaaigeneratedvideodetection, wang2025scalinglawsdeepfakedetection}. 

\subsubsection{Implementation Details}
We implement \modelname\ with GPT-4o~\cite{gpt-4o} as the vision–language backbone for the Hierarchical Perceptual Solver. To alleviate bias and failure modes from single-model reliance, we adopt Gemini-2.5-Pro~\cite{comanici2025gemini} as the backbone for the Self-Reflective Verifier. For video processing, we uniformly sample 8 frames as model input. In the Forensic Trace Extraction stage (Sec.~3.1), \texttt{grounding-dino-base}~\cite{liu2024grounding} is used for coarse localization, followed by SAM 2 initialized with the pre-trained {\nolinkurl{sam2.1_hiera_base_plus}} weights~\cite{ravi2024sam} for pixel-wise masking. The forensic toolbox $T = \{f_{\text{o}}, f_{\text{d}}, f_{\text{a}}, f_{\text{p}}\}$ integrates RAFT~\cite{teed2020raftrecurrentallpairsfield}, Depth Anything V2~\cite{yang2024depthv2}, DINOv2~\cite{oquab2023dinov2}, and D3~\cite{zheng2025d3trainingfreeaigeneratedvideo} to extract forensic cues from motion, depth, appearance, and pixel-level cues. All the tools are optimized on the \texttt{GenVideo} training split following the default configurations, while D3~\cite{zheng2025d3trainingfreeaigeneratedvideo} and Depth Anything V2~\cite{yang2024depthv2} are adapted and fine-tuned for synthesized video detection. In Eq.~\ref{eq6}, we set $\gamma=0.5$ and $L=3$.

\begin{table}[t]
\centering
\caption{
\textbf{Model comparison across SafeVid and other existing benchmarks.} \emph{T}, \emph{F}, \emph{O}, and \emph{P} denote \emph{Trainable}, \emph{Training-free}, \emph{Open-source}, and \emph{Proprietary} methods.
For evaluation efficiency, BusterX++~\cite{wen2025busterxpp}, Skyra-RL~\cite{li2025skyra}, and the generalist models are evaluated on randomly sampled 30\% subsets of each benchmark, and the human evaluators are assessed on 10\% randomly sampled subsets. In all cases, stratified sampling is applied to preserve the real/synth ratio. $^\dag$ indicates models pre-trained on \texttt{GenVideo} training set. Qwen3-VL and InternVL3.5 refer to Qwen3-VL-32B~\cite{Qwen3-VL} and InternVL3.5-38B~\cite{wang2025internvl3}, respectively. \best{Best} and \second{Second-best} are highlighted. \protect\colorbox{cyan!10}{Blue} highlights the best-performing baseline across both task-specific and generalist models.
}
\label{tab:main_results}

\renewcommand{\arraystretch}{1.20}
\setlength{\tabcolsep}{1.5pt} 
\scriptsize 
\resizebox{\linewidth}{!}{%
\begin{tabular}{l l *{16}{c}}
\toprule
\multirow{3}{*}[-1.75ex]{\textbf{Method}} &
\multirow{3}{*}[-1.75ex]{\textbf{Set}} &
\multicolumn{6}{c}{\textbf{\datasetsname\ (Ours)}} &
\multicolumn{8}{c}{\textbf{Existing Benchmarks}} &
\multicolumn{2}{c}{\textbf{Overall}} \\
\cmidrule(lr){3-8}\cmidrule(lr){9-16}\cmidrule(lr){17-18}

& &
\multicolumn{2}{c}{\textbf{ID}} &
\multicolumn{2}{c}{\textbf{OOD}} &
\multicolumn{2}{c}{\textbf{Avg.}} &
\multicolumn{2}{c}{\textbf{GenVideo}} &
\multicolumn{2}{c}{\textbf{DVF}} &
\multicolumn{2}{c}{\textbf{LOKI}} &
\multicolumn{2}{c}{\textbf{GenVidBench}} &
\multicolumn{2}{c}{\textbf{Avg.}} \\
\cmidrule(lr){3-4}\cmidrule(lr){5-6}\cmidrule(lr){7-8}
\cmidrule(lr){9-10}\cmidrule(lr){11-12}\cmidrule(lr){13-14}\cmidrule(lr){15-16}
\cmidrule(lr){17-18}

& &
\textbf{Acc} & \textbf{F1} &
\textbf{Acc} & \textbf{F1} &
\textbf{Acc} & \textbf{F1} &
\textbf{Acc} & \textbf{F1} &
\textbf{Acc} & \textbf{F1} &
\textbf{Acc} & \textbf{F1} &
\textbf{Acc} & \textbf{F1} &
\textbf{Acc} & \textbf{F1} \\
\midrule

\multicolumn{2}{l}{{Human}} &
{77.0} & {85.4} &
{80.9} & {80.0} &
{79.2} & {82.5} &
- & - & - & - & - & - & - & - &
{79.2} & {82.5} \\
\midrule

\multicolumn{18}{c}{\cellcolor{gray!10}\textit{\textbf{Task-Specific Models}}} \\
NPR~\cite{npr}  & \textit{T} &
71.6 & 75.5 &
62.8 & \cellcolor{cyan!10}{63.3} &
65.0 & \cellcolor{cyan!10}{66.6} &
85.3$^\dag$ & 71.6$^\dag$  & 
54.6 & 54.7 &
57.2 & 63.8 &
53.0 & 60.8 &
63.0 & 63.5 \\
FTCN~\cite{ftcn} & \textit{{T}} &
\cellcolor{cyan!10}{82.3} & \cellcolor{cyan!10}{82.5} &
{41.6} & {44.4} &
{51.6} & {53.5} &
\cellcolor{cyan!10}{95.6$^\dag$} & \cellcolor{cyan!10}{95.0$^\dag$} & 
{83.4} & {83.7} &
{68.1} & {74.3} &
{65.3} & {72.4} &
{72.8} & {75.8} \\

TALL~\cite{xu2023tall} & \textit{{T}} &
{70.4} & {74.1} &
{52.2} & {57.7} &
{56.7} & {61.7} &
{91.7}$^\dag$ & {90.5}$^\dag$ & 
{71.6} & {70.8} &
{65.9} & {72.8} &
{65.0} & {72.6} &
{70.2} & {73.7} \\
XCLIP~\cite{ni2022expanding}  & \textit{T} &
73.2 & 70.5 &
39.9 & 35.0 &
48.0 & 43.6 &
85.7$^\dag$ & 76.6$^\dag$ & 
82.1 & 82.2 &
66.0 & 69.7 &
62.7 & 69.6 &
70.0 & 71.1 \\
MINTIME~\cite{coccomini2024mintime} & \textit{{T}} &
{78.8} & {79.3} &
{45.5} & {44.2} &
{53.7} & {53.1} &
{95.3}$^\dag$ & {94.6}$^\dag$ & 
{83.6} & {83.9} &
{68.7} & {73.2} &
{64.8} & {71.8} &
{73.2} & {75.3} \\
DeMamba~\cite{chen2024demambaaigeneratedvideodetection} & \textit{T} &
{78.8} & {80.3} &
43.0 & 48.5 &
51.8 & 56.1 &
{94.4}$^\dag$ & {90.2}$^\dag$  &
\cellcolor{cyan!10}{90.1} & \cellcolor{cyan!10}{90.9} &
69.0 & 73.8 &
\cellcolor{cyan!10}{71.4} & \cellcolor{cyan!10}{78.3} &
\cellcolor{cyan!10}{75.5} & \cellcolor{cyan!10}{78.7} \\

RestraV~\cite{interno2025aigeneratedvideodetectionperceptual}  & \textit{T} &
36.1 & 49.7 &
46.5 & 61.0 &
43.9 & 58.4 &
73.9$^\dag$  & 72.9$^\dag$  &
33.3 & 47.4 &
51.9 & 66.5 &
61.8 & 76.1 &
53.0 & 64.3 \\

NSG-VD~\cite{zhang2025physics}& \textit{T} &
52.4 & 16.8 &
51.7 & 13.4 &
51.9 & 14.3 &
94.0$^\dag$ & {94.2}$^\dag$ &
68.0 & 43.4 &
32.7 & 3.2 &
67.0 & 52.6 &
62.7 & 41.5 \\

BusterX++~\cite{wen2025busterxpp} & \textit{T} &
69.7 & 68.8 &
\cellcolor{cyan!10}{64.7} & 55.2 &
\cellcolor{cyan!10}{66.8} & 61.7 &
76.1 & 85.0 &
70.4 & 66.6 &
\cellcolor{cyan!10}{70.4} & \cellcolor{cyan!10}{75.1} &
51.6 & 59.1 &
67.1 & 69.5 \\

Skyra-RL~\cite{li2025skyra} & \textit{T} &
63.3 & 43.5 & 
56.2 & 24.1 &
59.3 & 33.0 &
71.8 & 59.0 &
49.2 & 21.7 &
45.6 & 33.7 &
25.8 & 13.5 &
50.3 & 32.2 \\

D3~\cite{zheng2025d3trainingfreeaigeneratedvideo} & \textit{F} &
42.9 & 20.9 &
49.4 & 40.1 &
46.6 & 31.9 &
12.7 & 2.5 &
46.5 & 24.3 &
48.0 & 26.4 &
34.2 & 34.2 &
37.6 & 23.9 \\

\midrule

\multicolumn{18}{c}{\cellcolor{gray!10}\textit{\textbf{Generalist Models}}} \\
LLaVA-Video~\cite{zhang2024llavanextvideo} & \textit{O} &
35.9 & 19.2 &
44.7 & 28.3 &
41.0 & 24.3 &
48.3 & 62.4 &
56.6 & 55.5 &
46.7 & 48.9 &
53.0 & 64.8 &
49.1 & 51.2 \\

VideoLLaMA3~\cite{damonlpsg2025videollama3} & \textit{O} &
36.2 & 11.8 &
48.9 & 16.9 &
43.4 & 14.5 &
55.2 & 68.8 &
58.3 & 46.5 &
49.2 & 42.2 &
48.4 & 56.6 &
50.9 & 45.7 \\

InternVL3.5~\cite{wang2025internvl3} & \textit{O} &
42.9 & 18.1 &
52.7 & 19.7 &
48.5 & 19.0 &
58.5 & 70.9 &
61.2 & 50.1 &
56.3 & 55.2 &
47.3 & 54.3 &
54.4 & 49.9 \\

Qwen3-VL~\cite{Qwen3-VL} & \textit{O} &
62.0 & 60.2 &
62.1 & 47.2 &
62.1 & 53.7 &
78.4 & 86.7 &
75.0 & 73.2 &
70.0 & 74.8 &
58.3 & 67.2 &
68.8 & 71.1 \\

Video-R1~\cite{feng2025videor} & \textit{O} &
42.6 & 23.3 &
49.3 & 25.7 &
46.4 & 24.6 &
44.7 & 57.3 &
55.8 & 42.8 &
49.5 & 44.0 &
42.9 & 48.6 &
47.9 & 43.5 \\

Video-RFT~\cite{wang2025videorft} & \textit{O} &
43.3 & 20.2 &
49.1 & 22.0 &
46.6 & 21.2 &
45.8 & 58.5 &
55.0 & 39.6 &
48.0 & 40.4 &
42.4 & 47.8 &
47.5 & 41.5 \\

VideoChat-R1~\cite{li2025videochatr1} & \textit{O} &
44.0 & 24.8 &
53.4 & 25.9 &
49.4 & 25.4 &
56.1 & 69.0 &
60.2 & 48.6 &
50.9 & 45.3 &
45.3 & 52.2 &
52.4 & 48.1 \\

GPT-4o~\cite{gpt-4o} & \textit{P} &
65.9 & 70.1 &
63.7 & 52.0 &
64.6 & 60.5 &
77.6 & 86.0 &
74.5 & 75.6 &
55.8 & 66.4 &
61.5 & 68.8 &
66.8 & 71.5 \\

Gemini-2.5-Pro~\cite{comanici2025gemini} & \textit{P} &
64.3 & 59.0 &
58.4 & 39.6 &
61.0 & 48.9 &
83.2 & 89.8 &
76.8 & 78.9 &
59.6 & 70.4 &
54.6 & 60.9 &
67.1 & 69.8 \\

\midrule

\multicolumn{18}{c}{\cellcolor{gray!10}\textit{\textbf{Agentic Frameworks}}} \\

\multicolumn{2}{l}{\textbf{\modelname {\tiny(Qwen3-VL)} }} &
\second{92.2} & \second{90.0} &
\second{76.2} & \second{71.2} &
\second{82.9} & \second{79.7} &
\second{97.2} & \second{95.1} &
\second{96.3} & \best{93.2} &
\best{84.7} & \best{83.8} &
\second{80.6} & \second{78.7} &
\second{88.3} & \second{86.1} \\

\multicolumn{2}{l}{\textbf{\modelname {\tiny(GPT-4o)}}} &
\best{93.0} & \best{90.7} &
\best{80.1} & \best{72.8} &
\best{85.5} & \best{80.9} &
\best{97.9} & \best{95.6} &
\best{97.0} & \best{93.2} &
\second{81.7} & \second{77.7} &
\best{84.6} & \best{85.1} &
\best{89.3} & \best{86.5} \\

\bottomrule

\end{tabular}
}

\end{table}

\subsection{Main Comparison}

\subsubsection{Model Performance on SafeVid} As shown in Table~\ref{tab:main_results}, task-specific detectors achieve strong ID performance but exhibit severe degradation on OOD splits. Several video-specific and training-free methods perform close to random results, revealing the challenge of extracting reliable cues on SafeVid. Generalist models show modest performance on both splits, reflecting limited adaptation to AI-generated video detection. {\modelname} outperforms all baselines on ID and OOD sets, improving average Acc and F1 by +18.7\% over BusterX++~\cite{wen2025busterxpp} and +14.3\% over NPR~\cite{npr}. To further demonstrate the framework's versatility, we implement Qwen3-VL-32B~\cite{Qwen3-VL} as an open-source backbone for both the Solver and the Verifier. The consistent performance improvement across proprietary and open-source models validates the general effectiveness of our approach. Notably, \modelname\ achieves performance comparable to human evaluators {as reported in “Human” of Table~\ref{tab:main_results}}; however, the remaining OOD gap underscores the forensic difficulty of OOD videos.

\begin{table*}[t]
\centering

\caption{\textbf{Ablation study of \modelname\ across SafeVid and other existing benchmarks.} \textbf{SHGR}: Social Risk-Aware Hypothesis Generation and Routing, 
\textbf{VC}: Video Cropping, 
\textbf{FTE}: Forensic Trace Extraction, \textbf{SRV}: Self-Reflective Verifier.}
\label{tab:ablation_stages}

\renewcommand{\arraystretch}{1.0}
\setlength{\tabcolsep}{2.5pt} 
\scriptsize 

\resizebox{\linewidth}{!}{%
\begin{tabular}{l | cc | cc | cc | cc | cc | cc}
\toprule
\multirow{3}{*}[-1.75ex]{\textbf{Variant}} &
\multicolumn{2}{c|}{\textbf{\datasetsname}} &
\multicolumn{8}{c|}{\textbf{Existing Benchmarks}} &
\multicolumn{2}{c}{\textbf{Overall}} \\
\cmidrule(lr){2-3}\cmidrule(lr){4-11}\cmidrule(lr){12-13}

& \multicolumn{2}{c|}{\textbf{Avg.}} &
\multicolumn{2}{c|}{\textbf{GenVideo}} &
\multicolumn{2}{c|}{\textbf{DVF}} &
\multicolumn{2}{c|}{\textbf{LOKI}} &
\multicolumn{2}{c|}{\textbf{GenVidBench}} &
\multicolumn{2}{c}{\textbf{Avg.}} \\
\cmidrule(lr){2-3}\cmidrule(lr){4-5}\cmidrule(lr){6-7}\cmidrule(lr){8-9}\cmidrule(lr){10-11}\cmidrule(lr){12-13}

& \textbf{Acc} & \textbf{F1} & \textbf{Acc} & \textbf{F1} & \textbf{Acc} & \textbf{F1} & \textbf{Acc} & \textbf{F1} & \textbf{Acc} & \textbf{F1} & \textbf{Acc} & \textbf{F1} \\
\midrule

Baseline (GPT-4o)
& 64.6 & 60.5 & 77.6 & 86.0 & 74.5 & 75.6 & 55.8 & 66.4 & 61.5 & 68.8 & 66.8 & 71.5 \\
\hline
w/ SHGR
& 66.0 & 59.1 & 77.7 & 87.9 & 75.0 & 81.8 & 59.9 & 68.4 & 63.8 & 71.1 & 68.5 & 73.7 \\

w/ SHGR + VC
& 67.5 & 71.0 & 84.3 & 89.4 & 80.9 & 84.2 & 64.4 & 72.5 & 65.0 & 72.4 & 72.4 & 77.9 \\

w/ SHGR + VC + SRV
& 74.5 & 71.8 & 92.3 & 95.5 & 91.6 & 87.9 & 77.6 & 74.8 & 73.1 & 71.6 & 81.8 & 80.3 \\

w/ SHGR + VC + FTE
& 79.0 & 76.0 & 93.0 & 92.6 & 94.0 & 90.5 & 79.3 & 70.0 & 75.4 & 74.0 & 84.1 & 80.6 \\

\midrule
\textbf{\modelname (Ours)}
& \textbf{85.5} & \textbf{80.9} & \textbf{97.9} & \textbf{95.6} & \textbf{97.0} & \textbf{93.2} & \textbf{81.7} & \textbf{77.7} & \textbf{84.6} & \textbf{85.1} & \textbf{89.3} & \textbf{86.5} \\

\bottomrule
\end{tabular}%
}
\end{table*}
\begin{table*}[t]
\centering
\caption{
    \textbf{Ablation study of the forensic toolbox across SafeVid and other existing benchmarks.} $f_{o}$, $f_{d}$, $f_{a}$, and $f_{p}$ denote the tools that extract optical flow, depth, appearance, and pixel-level information from the input video, respectively. 
}
\label{tab:ablation_tools}

\renewcommand{\arraystretch}{1.0}
\setlength{\tabcolsep}{3.5pt} 
\scriptsize 

\resizebox{\linewidth}{!}{%
\begin{tabular}{l | cc | cc | cc | cc | cc | cc}
\toprule
\multirow{3}{*}[-1.75ex]{\textbf{Variant}} &
\multicolumn{2}{c|}{\textbf{\datasetsname}} &
\multicolumn{8}{c|}{\textbf{Existing Benchmarks}} &
\multicolumn{2}{c}{\textbf{Overall}} \\
\cmidrule(lr){2-3}\cmidrule(lr){4-11}\cmidrule(lr){12-13}

& \multicolumn{2}{c|}{\textbf{Avg.}} &
\multicolumn{2}{c|}{\textbf{GenVideo}} &
\multicolumn{2}{c|}{\textbf{DVF}} &
\multicolumn{2}{c|}{\textbf{LOKI}} &
\multicolumn{2}{c|}{\textbf{GenVidBench}} &
\multicolumn{2}{c}{\textbf{Avg.}} \\
\cmidrule(lr){2-3}\cmidrule(lr){4-5}\cmidrule(lr){6-7}\cmidrule(lr){8-9}\cmidrule(lr){10-11}\cmidrule(lr){12-13}

& \textbf{Acc} & \textbf{F1} & \textbf{Acc} & \textbf{F1} & \textbf{Acc} & \textbf{F1} & \textbf{Acc} & \textbf{F1} & \textbf{Acc} & \textbf{F1} & \textbf{Acc} & \textbf{F1} \\
\midrule

w/o tool use
& 74.5 & 71.8 & 92.3 & 95.5 & 91.6 & 87.9 & 77.6 & 74.8 & 73.1 & 71.6 & 81.8 & 80.3 \\
\hline
w/ $f_{o}$ 
& 75.6 & 74.7 & 93.1 & 95.2 & 93.6 & 88.1 & 78.2 & 70.0 & 74.6 & 72.4 & 83.0 & 80.1 \\

w/ $f_{o}+f_{d}$ 
& 78.5 & 78.4 & 95.1 & 95.0 & 94.0 & 90.0 & 78.7 & 73.9 & 78.7 & 76.0 & 85.0 & 82.7 \\

w/ $f_{o}+f_{d}+f_{a}$ 
& 83.3 & 79.5 & 96.2 & 95.1 & 96.3 & 92.8 & 81.6 & 75.3 & 82.5 & 80.6 & 88.0 & 84.7 \\

\midrule

w/ $f_{o}+f_{d}+f_{a}+f_{p}$ 
& \textbf{85.5} & \textbf{80.9} & \textbf{97.9} & \textbf{95.6} & \textbf{97.0} & \textbf{93.2} & \textbf{81.7} & \textbf{77.7} & \textbf{84.6} & \textbf{85.1} & \textbf{89.3} & \textbf{86.5} \\
\bottomrule
\end{tabular}%
}
\end{table*}

\subsubsection{Model Performance on Other Existing Benchmarks} 
To evaluate cross-dataset generalization, we further assess \modelname\ on \texttt{GenVideo}, \texttt{DVF}, \texttt{LOKI}, and \texttt{GenVidBench} in Table~\ref{tab:main_results}. As the task-specific models within the toolbox are pre-trained on \texttt{GenVideo}, the 97.9\% accuracy on this dataset serves as an in-domain reference. Without domain-specific fine-tuning, \modelname\ achieves 97.0\% on \texttt{DVF}, 81.7\% on \texttt{LOKI}, and 84.6\% on \texttt{GenVidBench}, outperforming state-of-the-art methods (DeMamba~\cite{chen2024demambaaigeneratedvideodetection} and BusterX++~\cite{wen2025busterxpp}) by margins of +6.9\%, +11.7\%, and +13.2\% respectively, demonstrating strong cross-dataset robustness. The performance on \texttt{LOKI} and \texttt{GenVidBench} highlights the inherent forensic challenges posed by advanced generative models, which produce highly realistic videos devoid of stereotypical artifacts. In contrast, \texttt{DVF} comprises videos generated by earlier models, exhibiting salient and easily detectable visual inconsistencies. Under the \texttt{LOKI} benchmark, \modelname\ (Qwen3-VL~\cite{Qwen3-VL}) outperforms \modelname\ (GPT-4o~\cite{gpt-4o}), indicating effective utilization of backbone-specific visual priors under diverse artifact distributions.

\subsection{Ablation Study}
We conduct ablation studies to quantify the contribution of individual modules within \modelname. By default, the Solver utilizes GPT-4o~\cite{gpt-4o} and the Verifier employs Gemini-2.5-Pro~\cite{comanici2025gemini}. Across all variants, the evaluation protocol remains strictly constant, modifying exclusively the ablated components.

\subsubsection{Effect of the Hierarchical Perceptual Solver} 
Table~\ref{tab:ablation_stages} summarizes the impact of the \emph{Hierarchical Perceptual Solver} (HPS), comprising SHGR, VC, and FTE. Progressive integration of these components increases Avg. Acc from 66.8\% to 84.1\%. Specifically, SHGR yields an initial improvement, VC strengthens hypothesis-guided spatial grounding, and FTE introduces a substantial performance enhancement (+11.7\% Avg. Acc), underscoring the critical necessity of explicit forensic evidence. Overall, the HPS converts suspicious-region hypotheses into verifiable, localized traces of potential forgery.

\begin{figure*}[t]
    \centering
        \centering
        \includegraphics[width=1.0\linewidth]{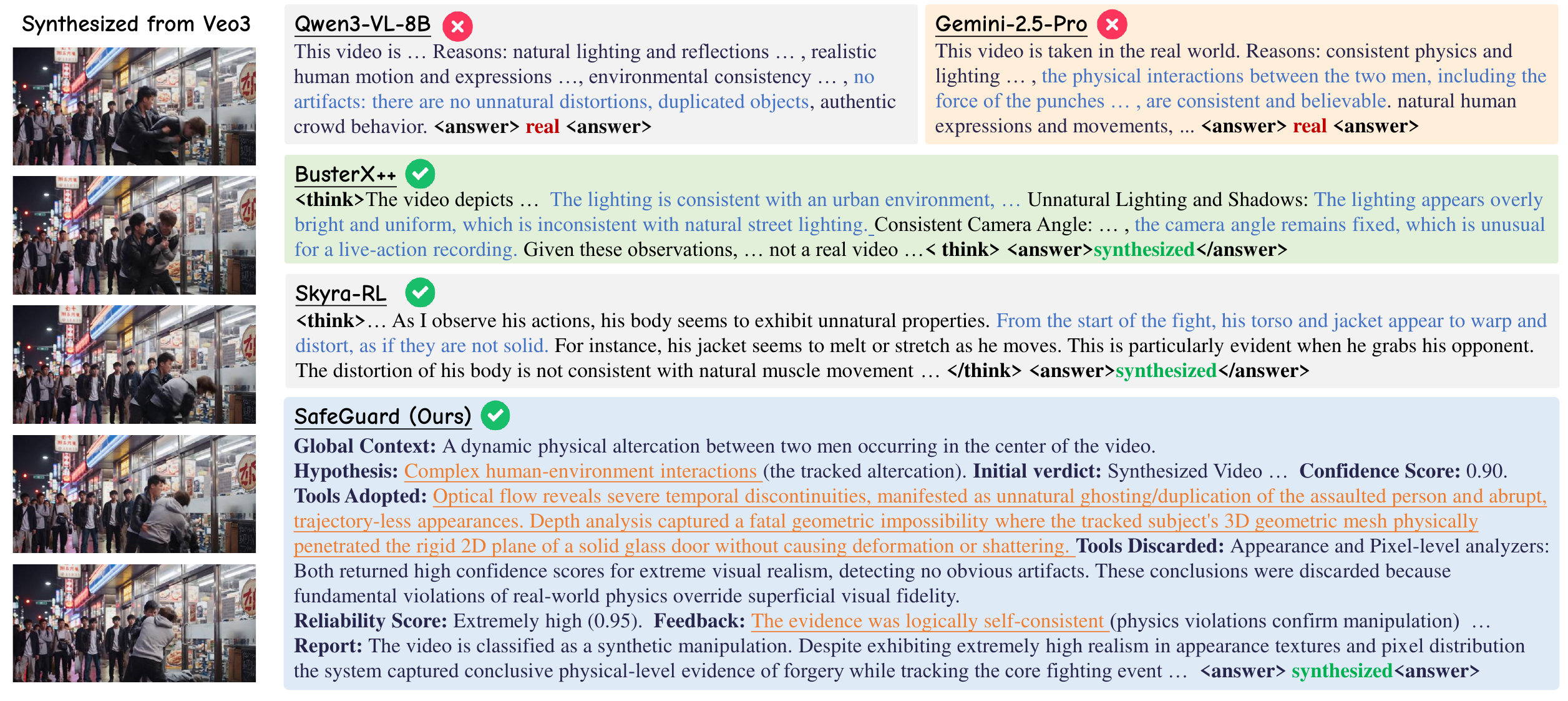}
        \caption{\textbf{A case study of \modelname\ and other existing models on SafeVid.} 
        The visual anomalies within the video manifest as repeating humanoid outlines during combat and figures that appear abruptly without plausible trajectories. 
        }
        \label{fig:case_study}
    \hfill
\end{figure*}

\subsubsection{Effect of the Self-Reflective Verifier}  
Table~\ref{tab:ablation_stages} demonstrates the efficacy of the \emph{Self-Reflective Verifier} (SRV), which enhances overall performance by +5.2\% Avg. Acc compared to the SHGR+VC+FTE. This consistent improvement across benchmarks indicates that raw forensic evidence extraction remains insufficient for reliable decision-making. Notably, activating the SRV without FTE yields performance degradation (81.8\% Avg. Acc), demonstrating that reflective reasoning cannot independently compensate for an absence of explicit forensic evidence. Rather, the SRV functions to enforce hypothesis–evidence consistency and mitigate reasoning discrepancies through closed-loop refinement.

\subsubsection{Effect of the Forensic Toolbox}
Table~\ref{tab:ablation_stages} indicates that integrating the FTE substantially improves performance over reasoning alone (+9.4\% Avg. Acc), further underscoring the necessity of grounded forensic evidence. As detailed in Table~\ref{tab:ablation_tools}, progressively aggregating task-specific models within the toolbox produces consistent accuracy improvement, culminating in 89.3\% Avg. Acc and 86.5\% Avg. F1. Specifically, the integration of depth and appearance models contributes significantly to these observed improvements.

\subsection{Case Study}
As shown in Fig.~\ref{fig:case_study}, the generalist models Qwen3-VL~\cite{Qwen3-VL} and Gemini-2.5-Pro~\cite{comanici2025gemini} over-rely on the realistic appearance and incorrectly predict \textit{real}. BusterX++~\cite{wen2025busterxpp} outputs the correct label but provides only a coarse, high-level explanation without identifying concrete forensic traces. Skyra-RL~\cite{li2025skyra} adopts a generic artifact-checklist reasoning approach, which is weakly grounded in the subtle, localized inconsistencies present in this high-fidelity video. In contrast, \modelname\ correctly predicts \textit{synthesized} and delivers a complete, reproducible reasoning pipeline, supplying traceable evidence at each step.

\section{Conclusion}

To address the scarcity of datasets covering socially sensitive threats and the \emph{Perception–Reasoning Gap} in existing detectors, we introduce \datasetsname, a benchmark for high-risk scenarios, and \modelname, a multi-agent forensic framework. Our analysis shows that task-specific detectors often fail on out-of-domain data, while general large vision-language models overlook subtle visual artifacts. \modelname\ mitigates these issues by combining forensic tools with semantic reasoning, providing a transparent, verifiable chain of evidence and robust support for high-stakes, socially sensitive forensic scenarios.

\section*{Acknowledgements}
This work was supported by the National Natural Science Foundation of China (NSFC) under Grant U22A2094 and Grant 62272435, and also supported by the Yangtze River Delta Science and Technology Innovation Community Joint Research (Basic Research) Project under Grant 2025CSJZN01600.

\newpage
%
%
\bibliographystyle{splncs04}
\bibliography{main}

\end{document}